\documentclass[10pt, conference, compsoconf]{IEEEtran}

\usepackage{cite}
\usepackage{amsmath,amssymb,amsfonts}
\usepackage{algorithmic}
\usepackage{graphicx}
\usepackage{textcomp}
\usepackage{xcolor}
\def\BibTeX{{\rm B\kern-.05em{\sc i\kern-.025em b}\kern-.08em
    T\kern-.1667em\lower.7ex\hbox{E}\kern-.125emX}}
\usepackage{balance}
\usepackage{leading}
\usepackage{subfig}
\usepackage{tabularx}
\usepackage{textcomp}
\usepackage{multirow}
\usepackage{framed}
\usepackage{listings}
\usepackage{float}
\usepackage{enumitem}
\usepackage{comment}

\input{macros}

\graphicspath{{../graphs/nsdi_plots/}}

\begin{document}


\title{Taming Resource Heterogeneity In Distributed ML Training With Dynamic Batching}


\author{\IEEEauthorblockN{Sahil Tyagi \hspace{2ex} Prateek Sharma} \linebreak
\IEEEauthorblockA{
\emph{Indiana University Bloomington} \\
\emph{\{styagi, prateeks\}@iu.edu}}
\vspace*{-25pt}
}


\maketitle
\thispagestyle{empty}
\pagestyle{empty}

\begin{abstract}
  Current techniques and systems for distributed model training mostly assume that clusters are comprised of homogeneous servers with a constant resource availability. 
However, cluster heterogeneity is pervasive in computing infrastructure, and is a fundamental characteristic of low-cost transient resources (such as EC2 spot instances). 
In this paper, we develop a dynamic batching technique for distributed data-parallel training that adjusts the mini-batch sizes on each worker based on its resource availability and throughput. 
Our mini-batch controller seeks to equalize iteration times on all workers, and facilitates training on clusters comprised of servers with different amounts of CPU and GPU resources. 
This variable mini-batch technique uses proportional control and ideas
from PID controllers to find stable mini-batch sizes.
Our empirical evaluation shows that dynamic batching can reduce model training times by more than $4\times$ on heterogeneous clusters.

\end{abstract}





\vspace*{\subsecspace}
\section{Introduction}
\vspace*{\subsecspace}


Distributed training of machine learning models by using large clusters of servers is a popular technique to decrease the model training time. 
Techniques and system-architectures for distributed ML training such as  Stochastic Gradient Descent (SGD) and parameter servers are widely used to train in data centers and cloud platforms to provide reasonable parallel speedup.

However, current techniques and systems for distributed model training mostly assume that the workers (i.e., the servers) will all have the same performance and  resource configuration, i.e., will be homogeneous. 
However, virtual clusters in data centers and especially clouds \emph{do not} always exhibit this resource homogeneity. 
The performance of different workers can be affected due to performance interference with co-located applications; workers may be throttled by the cloud or data center provider; or the cluster may have servers with vastly different resource configurations.

This \emph{resource heterogeneity} is a key characteristic of cloud-based applications, and distributed ML model training must be able to tolerate and perform well even in heterogeneous environments. 
However, heterogeneity presents many fundamental challenges to distributed training: synchronous model updates result in stragglers causing poor parallel efficiency, and asynchronous updates result in gradient and model staleness causing poor statistical efficiency~\cite{recht2011hogwild}.

In this paper, we address the challenges of distributed ML training in heterogeneous environments. 
Our goal is to make model training ``omnivorous'', and be able to run efficiently on dynamic and heterogeneous cluster configurations in shared data center and cloud environments. 
Our key insight is that having variable, instead of uniform mini-batch sizes on different workers, is a simple yet powerful technique for alleviating many of the performance degradation problems in heterogeneous environments. 

Our dynamic batch sizing mechanism adjusts the mini-batch size on each worker based on the worker's throughput, by using a proportional-controller~\cite{pid-wiki}  that minimizes the differences in the workers' iteration times.
This dynamic batch sizing technique permits training on clusters made up of servers with vastly different resource configurations; and on clusters with dynamic resource availability due to resource elasticity, over-commitment, or preemption. 
The technique enables us to train models efficiently on clusters comprising of servers with different CPUs and GPUs, which is a key differentiator from prior work in heterogeneous distributed training~\cite{jiang_heterogeneity-aware_2017,luo_hop:_2019} that instead focuses on random worker slowdowns. 
The prior work has shown that even small random slowdowns can result in the training times increase by an order of magnitude.
This is only exacerbated with the systemic heterogeneity that we aim to alleviate.

The dynamic batching mechanism is able to reduce stragglers in Bulk Synchronous Parallel (BSP) training, and is designed as zero-configuration, black-box approach that can effectively work with different training, model, and resource configurations. 
Our approach allows distributed training on clusters with dynamic resource availability that are ubiquitous in cloud environments. 
%
By mitigating the performance degradation due to heterogeneity, our contributions enable low-cost training on heterogeneous collections of transient cloud servers such as EC2 spot instances~\cite{ec2-spot} and Google Preemptible VMs~\cite{preemptible-vm}, that are up to $10\times$ cheaper than conventional cloud servers. 
%
%
We implement our dynamic batching mechanism and policies in TensorFlow, and make the following contributions:

\begin{enumerate}
\item We develop a dynamic batching mechanism for data-parallel training that continuously balances the load between workers by assigning them different sized mini-batches based on their throughput. Our proportional-control based technique reduces stragglers in BSP, and allows the mixing of CPU and GPU servers. It is able to ameliorate both static as well as dynamic heterogeneity. 


\item We implement all our mechanisms and policies in TensorFlow using the estimator API, which allows most models to directly run in heterogeneous environments without any modifications.

\item We conduct a large scale study of training performance in various static and dynamic heterogeneity environments using popular ML workloads. Our techniques can reduce the training times by as much as $4\times$ compared to existing uniform-batching. 
  
\end{enumerate}

\vspace*{\subsecspace}
\section{Background \& Motivation}


\subsection{Heterogeneity in Data Centers and Clouds}

Resource heterogeneity is pervasive in modern data centers and clouds. 
In cloud environments, applications are often deployed on clusters  composed of servers (i.e, VMs) of different resource capacities and sizes.
This \emph{static heterogeneity} is \emph{necessary} for effectively using low-cost transient servers such as Amazon EC2 spot instances~\cite{ec2-spot}, Google Preemptible VMs~\cite{preemptible-vm}, etc.

Since distributed model training is highly computationally intensive, using low-cost transient VMs or low-priority data center resources is a key technique for reducing training costs~\cite{li_speeding_2019, tian_gpu_icdcs}.
Transient VMs can be unilaterally preempted by the cloud provider, which are akin to fail-stop failures.
Distributed applications that can tolerate a failure of a (small) subset of their servers failing can benefit greatly from running on VMs of different sizes. 
Past work on transient computing~\cite{exosphere,flint} has found that transient VMs of different sizes are usually \emph{uncorrelated} in their preemptions, and this diversification significantly reduces the risk of all the VMs preempted at the same time. 
Thus distributed training needs to be ``omnivorous'', capable of using different types of low-cost low-priority servers and cannot assume homogeneous clusters with constant resource availability.

\vspace*{\subsecspace}
\subsection{Distributed Training}
\vspace*{\subsecspace}

Training of machine learning models entails learning the model parameters (a.k.a weights) of a given model (such as a deep neural network) over an input training dataset. 
This process is typically done through an iterative-convergent process that gradually minimizes some loss function of the model over the dataset, by using an optimization technique such as Stochastic Gradient Descent (SGD)~\cite{sgd}.

Since ML training is highly compute intensive, parallelizing it using computational accelerators such as GPUs and TPUs via distributed training is vital~\cite{torsten_demystifying_2018,mayer_scalable_2019}. 
In distributed training, multiple \emph{workers} participate to iteratively refine the model.
A common parallelization approach is \emph{data-parallelism}, where training is launched on multiple workers, and each worker learns and updates the model parameters by processing a small batch of the training data~\cite{dean2012large}. 
Each iteration comprises of computing model updates to the previous model parameters, and sharing the updates with other workers to form a new global model. 
Training a popular image recognition model such as ResNet~\cite{szegedy2017inception} requires tens of thousands of iterations until the model's error converges to a low-enough target.

Conventionally, workers send their updates to a smaller number of parameter servers that apply the updates and compute an ``averaged'' model that is broadcasted to workers before the next iteration~\cite{li_scaling_2014}. 
Concretely, the learning process involves iteratively computing the model parameters over $K$ workers, each processing a mini-batch of $b$ items at iteration $t$ and computing the gradient $\nabla f(\mathbf{x}_{k, t})$.
The gradients from all the workers are then collected and averaged by the parameter servers, and the update rule for the model parameters $\mathbf{x}$ is given by :     
\begin{equation}
  \mathbf{x}_{t+1} = \mathbf{x}_{t} - \eta \dfrac{1}{K} \dfrac{1}{b} \sum_{k=1}^{k=K}{\nabla f(\mathbf{x}_{k, t})},
    \label{eq:sgd}
\end{equation}
where $\eta$ is the learning rate parameter which is one of the ``hyperparameters'' of the model that is  found through empirical search techniques.

%

\vspace*{\subsecspace}
\subsection{Training Challenges in Heterogeneous Environments}
\vspace*{\subsecspace}

\begin{figure}[t]
  \centering
  \includegraphics[width=0.27\textwidth]{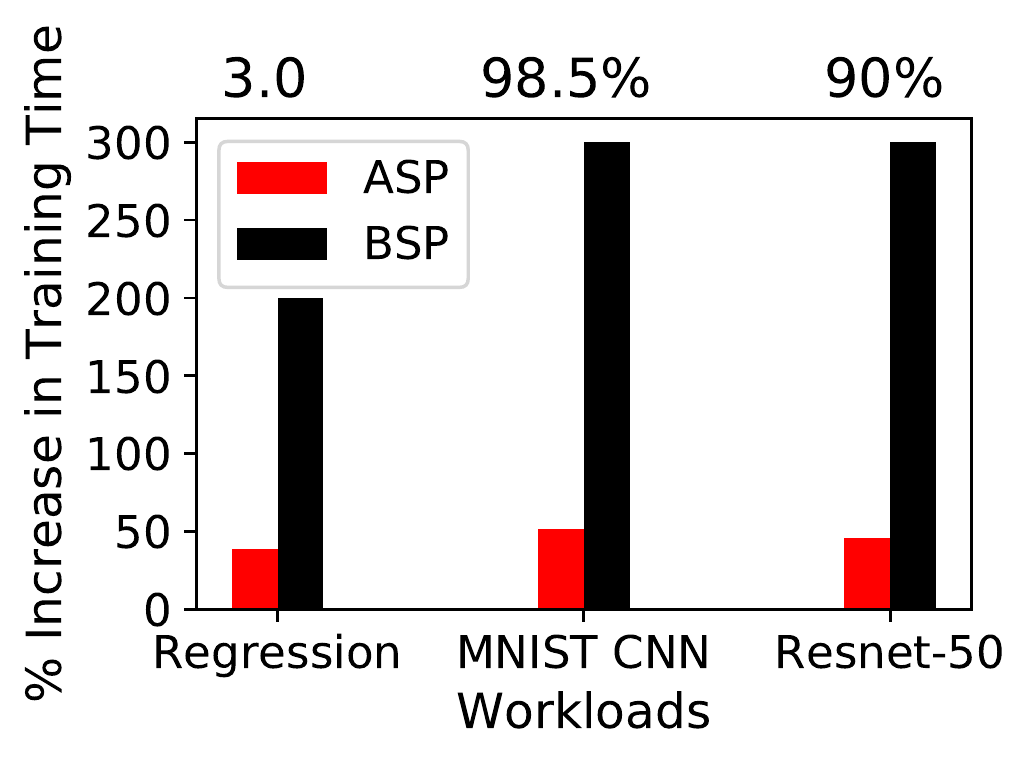}
   \vspace*{\captionspace}
   \caption{Increase in the total training time compared to a homogeneous cluster for three popular ML workloads. Both the homogenous and heterogenous clusters have the same total amount of computing resources.}
  \label{fig:baseline-workloads}
  \vspace*{\captionspace}
\end{figure}

If the computing capacities of the workers is not uniform and constant, then data-parallel training suffers from severe performance degradation. 
The performance and model quality (i.e., model accuracy) of distributed training is \emph{highly} dependent on the communication and synchronization of the gradient updates to compute new model parameters. 
In bulk synchronous parallel (BSP) SGD, new model parameters are computed after gradients from \emph{all} workers have been received.
Even in homogeneous conditions, stragglers are an important concern in synchronous data-parallel training.
In heterogeneous environments, straggler workers with lower computational resources will take much longer to process their mini-batches. 
Thus, BSP suffers from poor parallel efficiency in heterogeneous environments because of stragglers that significantly increase the total training time. 

The performance impact of heterogeneity for distributed data parallel training in TensorFlow is shown in Figure~\ref{fig:baseline-workloads}. 
The figure compares the total wall-clock training time to a desired accuracy level of a homogenous vs.  heterogeneous cluster.
In all cases, the total amount of computing resources is held constant.
The homogenous cluster has three workers with 10 CPUs each, and the heterogeneous cluster has workers with (3,9,18) CPUs.
We see that heterogeneity can result in training times to increase by more than 300\% compared to the homogenous case for BSP, and 50\% for asynchronous (ASP) training. 

\vspace*{\subsecspace}
\section{Dynamic Mini-Batching} \label{sec:mech}
 \vspace*{\subsecspace}

In this section, we describe our dynamic batching mechanism and policies for distributed training. 
Our focus is on data-parallel training on heterogeneous clusters of data center or cloud servers.

\begin{figure}[t]
  \centering
  \includegraphics[width=0.4\textwidth]{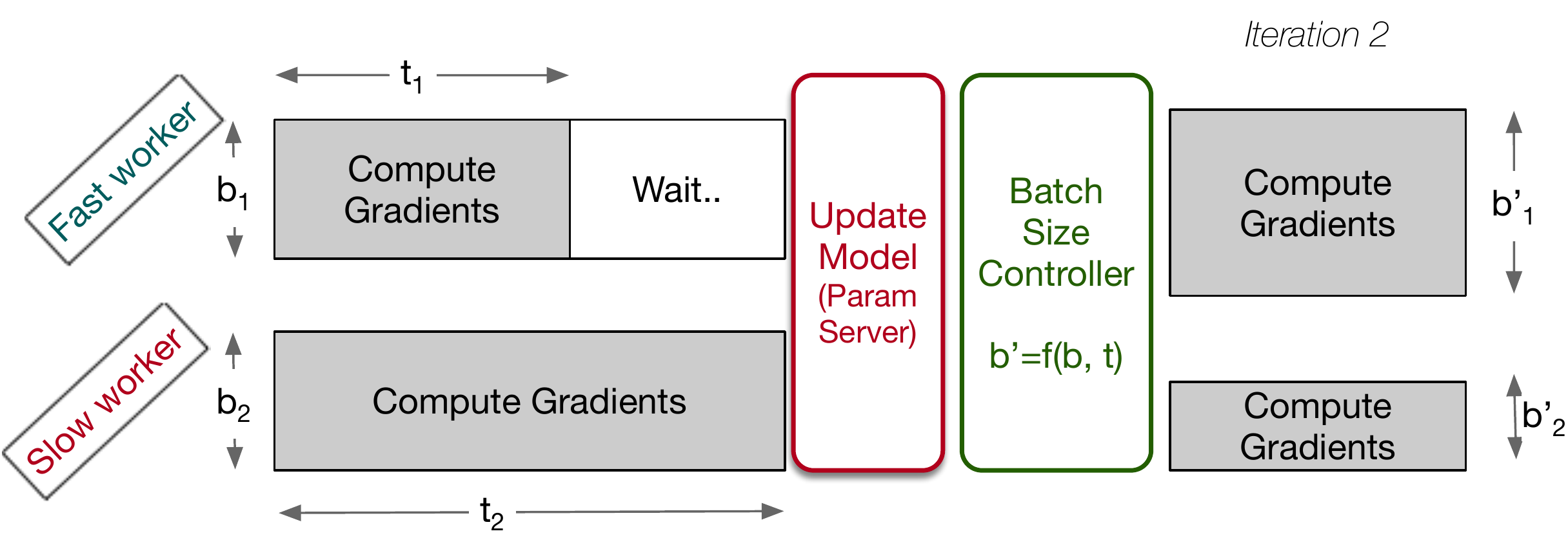}
  \vspace*{\captionspace}
  \caption{With variable batching, we can decrease the batch size on the slower worker, and increase the batch size on the larger worker, so that no worker ``waits'' for another.}
    \label{fig:overview}
  \vspace*{\myfigspace}
\end{figure}

\vspace*{\subsecspace}
\subsection{Key Idea: Variable Mini-Batch Sizes} \label{subsec:var-batch}
\vspace*{\subsecspace}

Conventional data-parallel training uses mini-batch SGD for distributing and parallelizing the model training process. 
This approach entails each worker processing a mini-batch of training samples independently and computing the gradients. 
The gradients are computed over the mini-batch of size $b$ by all the workers. 
Due to  resource heterogeneity, the mini-batch processing times (i.e., iteration times) across workers can be different. 
This results in stragglers in the case of BSP and staleness in the case of ASP---both of which cause a significant increase in the model training time to a desired accuracy level. 

The main insight is that the mini-batch sizes need not be uniform across workers---instead, the mini-batch sizes should be proportional to the server resource availability. 
This \emph{variable} batching allows workers to process different amount of data. 
The goal is to reduce the differences in the workers' iteration times to reduce stragglers and staleness. 
This is illustrated in Figure~\ref{fig:overview}, which shows the use of variable batching during the training process to adjust the worker batch sizes to minimize stragglers.

Such variable batching is compatible with distributed SGD---we assign a mini-batch size of $b_k$ to worker $k$. 
Because workers are processing different amounts of training data, their contribution in the training process is no longer uniform.
In conventional SGD, the gradients from all workers are averaged as per Equation~\ref{eq:sgd}. 
However, with variable batching, the gradients computed by workers with larger batch sizes need to be ``weighted'' more than the gradients computed using smaller batches.  
Thus we \emph{scale} the gradients computed by each worker based on its mini-batch size, and the final gradients are computed using a weighted average. 

We use linear gradient scaling: gradients on worker $k$ are multiplied by $\lambda_k$ such that $\lambda_k \propto b_k$. 
To maintain equivalence with conventional uniform batching, we also require that $\sum_k{\lambda_k} = 1$.
This yields $\lambda_k = \frac{b_k}{\sum_{i=1}^{i=K} b_i}$. 
The new weights for the next iteration are then computed by doing a weighted average of the gradients:\vspace*{-5pt}
\begin{align}
  \label{eq:wsgd2}
  g_{k, t} & = \lambda_k \nabla f(\mathbf{x}_{b_k, t}) \\
  \mathbf{x}_{t+1} & = \mathbf{x}_{t} - \frac{1}{K} \eta \sum_{k=1}^{k=K} g_{k,t},
\end{align}
where $\nabla f(\mathbf{x}_{b_k, t})$ is the gradient computed using mini-batch $b_k$ by worker $k$.
The weighted averaging is done by the parameter server, and preserves the convergence properties of SGD~\cite{ferdinand_anytime_2019}. 




Ideally, we want perfect load balancing, and assign mini-batch-sizes to workers such that all workers finish their iterations at the same time. 
Due to servers of different sizes and dynamic resource availability due to interference or over-commitment, the processing power on workers also varies.
In the next two subsections, we describe two approaches for assigning mini-batches to workers---a simple static open-loop allocation technique, and a closed-loop dynamic allocation that can respond to cluster resource dynamics.

\vspace*{\subsecspace}
\subsection{Static Mini-batch Allocation Policy}
\vspace*{\subsecspace}
\label{subsec:static-batch}

Instead of uniform mini-batches for all workers, our \emph{static assignment} policy computes mini-batch sizes proportional to the worker's computing power. 
Because model training is highly computation-bound, the throughput of workers is proportional to the available CPU and GPU resources. 
Thus given a heterogeneous cluster of $K$ workers, we want $b_k \propto X_k$, where $X_k$ is the throughput (i.e., training samples processed per second) of server $k$. 

We seek to maintain the initial average mini-batch size, $b_0$ that is  provided for a given ML model. 
We then allocate the mini-batches to different workers such that the batch sizes are proportional the worker throughput, and the global batch size is maintained:  $b_k = \frac{b_0 X_k}{\sum_i X_i}$ .
This ensures that $\sum_i b_i = K b_0$, where $b_0$ corresponds to the conventional uniform mini-batch size. 
Importantly, this approach keeps the total global batch size constant, and invariant to variable batching. 

\begin{figure}[t]
  \vspace*{\captionspace}
  \subfloat[Uniform batching. \label{fig:pdf-uniform}]
  {\includegraphics[width=0.24\textwidth]{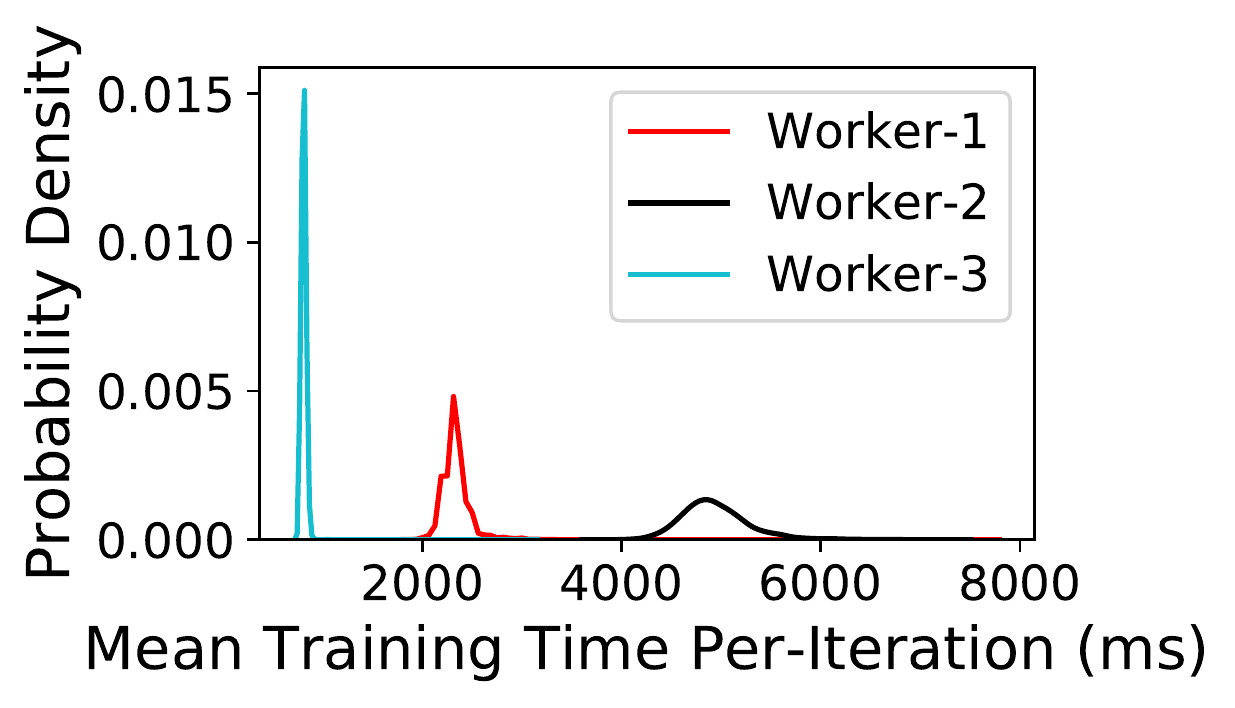}}
  \hfill \hspace*{-10pt}
  \subfloat[Variable batching. \label{fig:pdf-var}]
  {\includegraphics[width=0.24\textwidth]{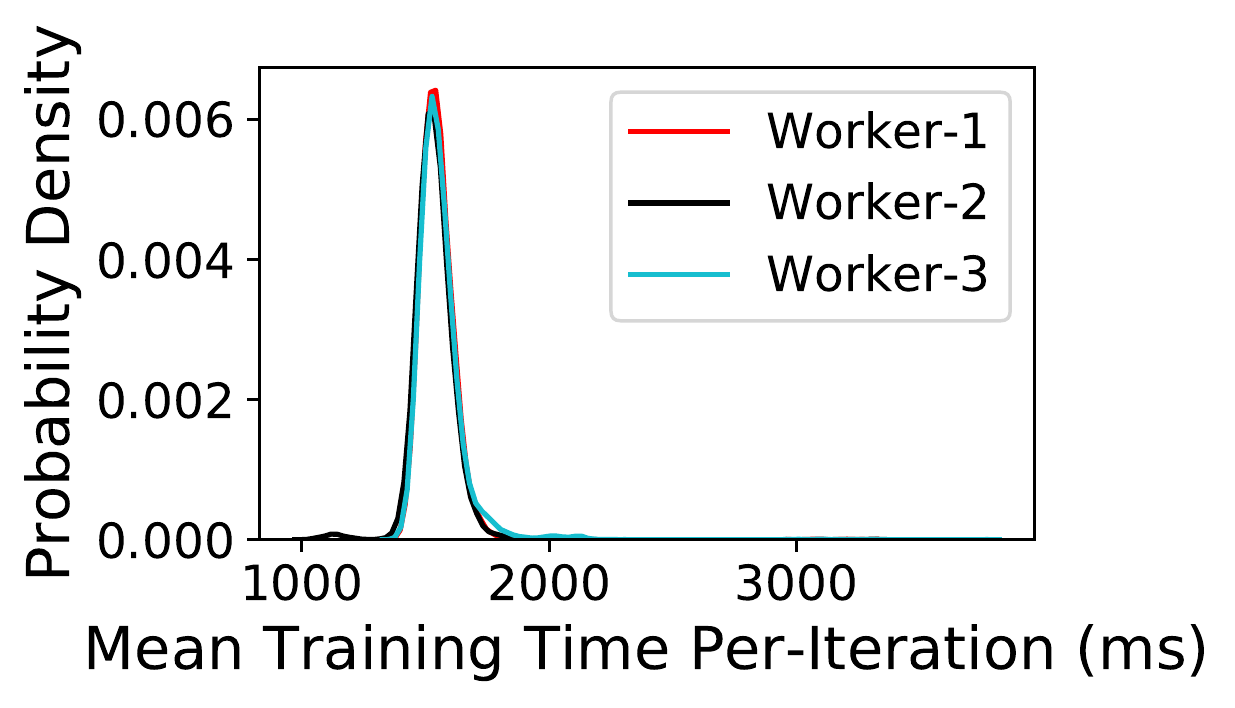}}
  \hfill
  \vspace*{\captionspace}
  \caption{Frequency distributions of iteration times across workers in a heterogeneous cluster. Worker 3 is 3x larger than worker 1, which is 2x larger than worker 2. Variable batching ensures that the iteration times across workers are similar.} 
  \label{fig:time-dist}
  \vspace*{\myfigspace}
\end{figure}

Static mini-batch allocation seeks to ``equalize'' the iteration times on different workers, as illustrated in Figure~\ref{fig:time-dist}, which 
 shows the distribution of iteration times for ResNet-50 (BSP)\footnotemark, with three workers in a heterogeneous cluster with $(3,5,12)$ CPU cores respectively. 
With  uniform batching in Figure~\ref{fig:pdf-uniform}, the iteration times for the workers are different, due to the differences in their processing powers. 
In contrast, with the variable mini-batching approach, the iteration times for all the workers have similar frequency distributions, as seen in Figure~\ref{fig:pdf-var}.

By reducing the gap between iteration times among workers, the variable batching technique can reduce stragglers in the case of BSP and thus the total training time in heterogeneous environments.
Unlike for BSP, our approach does not \emph{directly} address the root cause of slowdowns for ASP training.
With ASP, the slowdown is a result of the statistical inefficiency arising due to multiple factors, including gradient update staleness.
However the relation between staleness and training time is not as simple to model as the effect of stragglers on BSP~\cite{zhou_distributed_nodate,dutta_slow_2018}, and is not necessarily linear. 
Nevertheless, reducing the iteration gap allows us to ameliorate the staleness and improve the total training time even for ASP, albeit not as effectively as BSP. 


\footnotetext{We train ResNet with BSP for all the examples and figures in this section.}

\noindent \textbf{Estimating throughput.}
We can estimate the worker throughput required for the variable batch allocation based on the server's resource configuration. 
When workers are using only CPU resources, a simple way is to assign batch sizes proportional to the number of CPU cores. 
In case a distributed training job is running on both CPU and GPU servers,  we assign batch sizes proportional to the half-precision FLOPs (Floating Point Operations per Second) for each server. 
This a one-shot method that is black box and requires no adjustment, and is ``open-loop''.

However,throughput may not be exactly proportional to the server FLOPs. 
This error can cause imperfect load-balancing, and can result in sub-optimal mini-batching. Compared to the ideal batch sizes that equalize all iteration times, some workers may get smaller batches and wait for workers with larger batches. 
We can address this problem by \emph{dynamically} assigning mini-batch sizes, which we describe in the next subsection. 



\vspace*{\subsecspace}
\subsection{Proportional-Control Based Dynamic Policy}
\vspace*{\subsecspace}
\label{subsec:dyn-batch}

To mitigate stragglers and staleness, it is crucial for workers to finish processing their mini-batches simultaneously. 
In the previous static technique, the mini-batches were allocated based on the estimated relative throughput of different workers. 

This open-loop estimation, based on the hardware FLOPs, is not accurate in predicting the training throughput, in two major situations.
First, the training throughput depends on intra-worker parallel scaling characteristics governed by Amdahl's law.
Thus the observed throughput on large workers (with more CPUs) may be lower than what is indicated by their core counts. 
Second, many scenarios yield \emph{dynamic} resource availability, which the static mini-batching approach is ill-suited for.



Our \emph{dynamic} mini-batching technique is designed to handle throughput estimation errors, as well as handling dynamic resource availability due to server overcommittment or intermittent performance interference that lead to variable effective throughput on the affected workers. 
The key idea is to continuously adjust the mini-batch sizes on the workers. 
The goal is to equalize the iteration times among all the workers.
Let worker $k$ finish computing gradients for its mini-batch in time $t_k$.
Ideally, we want $t_i = t_j$ for all workers $i,j$.

The dynamic mini-batch adjustment uses a simple proportional-control approach to compute the mini-batch size of all workers. 
Since the goal is to equalize the batch processing times, the ``error'' is $\tau_k = t_k - \bar{t}$, where $\bar{t}$ is the average iteration time across all the workers. 
To minimize this error, the mini-batch size is updated by $\Delta(b)$ by the following proportional control rule: 
\begin{equation}
  \Delta(b_k) = -X_k \tau_k,
    \label{eq:dbatch1}
\end{equation}
where, $X_k$ is the throughput of worker $k$, which can be empirically determined as $X_k = b_k/t_k$.
The new batch size for iteration $i+1$ is computed as follows: 
\begin{equation}
    b_k^{i+1} = b_k^{i} + \Delta(b_k^{i}),
    \label{eq:dbatch2}
\end{equation}
Thus slower workers ($t > \bar{t}$) will have a positive error $\tau$, and their batch sizes will be decreased. Workers whose batch processing times are faster than average, are capable of handling a higher load, and will get a larger batch size after the dynamic adjustment. 
Simplifying the above two equations, we can compute the new batch size $b_k^{(1)}$ from the initial batch size $b_k^{(0)}$ as : $b_k^{(1)} = b_k^{(0)}\bar{t}/t_k$.

This policy essentially combines model-based and conventional black-box PID controllers.
Instead of using and tuning an arbitrary proportionality constant like in most PID controllers, we use the (estimated) throughput. 




\noindent \textbf{Initial mini-batch sizes.}
The dynamic mini-batching approach works with any initial batch size. 
By default, the initial batch sizes are allocated based on the throughput-based open-loop variable batching approach described in the previous subsection. 
In that case, any error in the throughput approximation (based on the CPU/GPU FLOPs) is corrected by the control mechanism. 

While a good starting point is desirable, it is not necessary.
The dynamic batching approach permits \emph{any} initial batch size allocation, with the caveat that the farther the initial batch size is from the ideal (i.e., throughput proportional), the larger number of batch adjustment steps are required to reach the equilibrium steady-state batch sizes.
For example, Figure~\ref{fig:dyn-timeline-smooth} shows the progress of the batch adjustment on three heterogeneous workers when all the workers are assigned the same initial batch size (which is sub-optimal).
We can see that the mini-batch sizes on the different workers converge to their stable throughput-proportional values after only two batch adjustments.
Thus, the dynamic batching technique is useful in situations where apriori throughput estimates are not be known. 




\subsubsection{Control stability}
\label{subsubsec:control-stability}

\begin{figure}[t]
  \centering
  \vspace*{\myfigspace}
  \subfloat[Batch sizes quickly converge within two adjustments. \label{fig:dyn-timeline-smooth}]
  {  \includegraphics[width=0.22\textwidth]{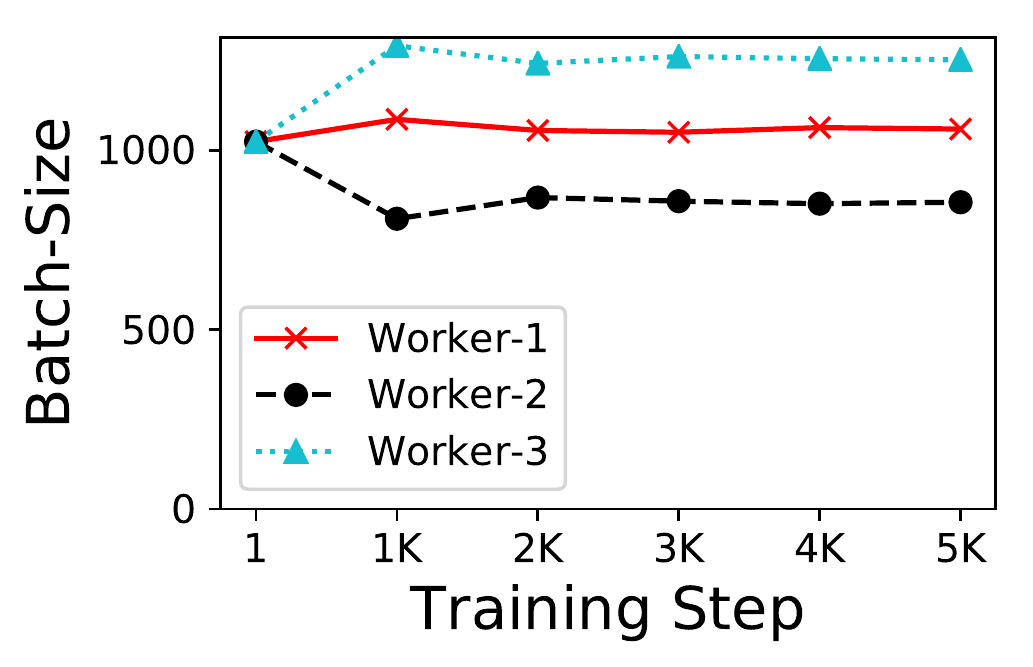}}
  \hfill 
  \subfloat[Oscillations in mini-batch sizes without dead-banding.\label{fig:ping-pong}]
{    \includegraphics[width=0.22\textwidth]{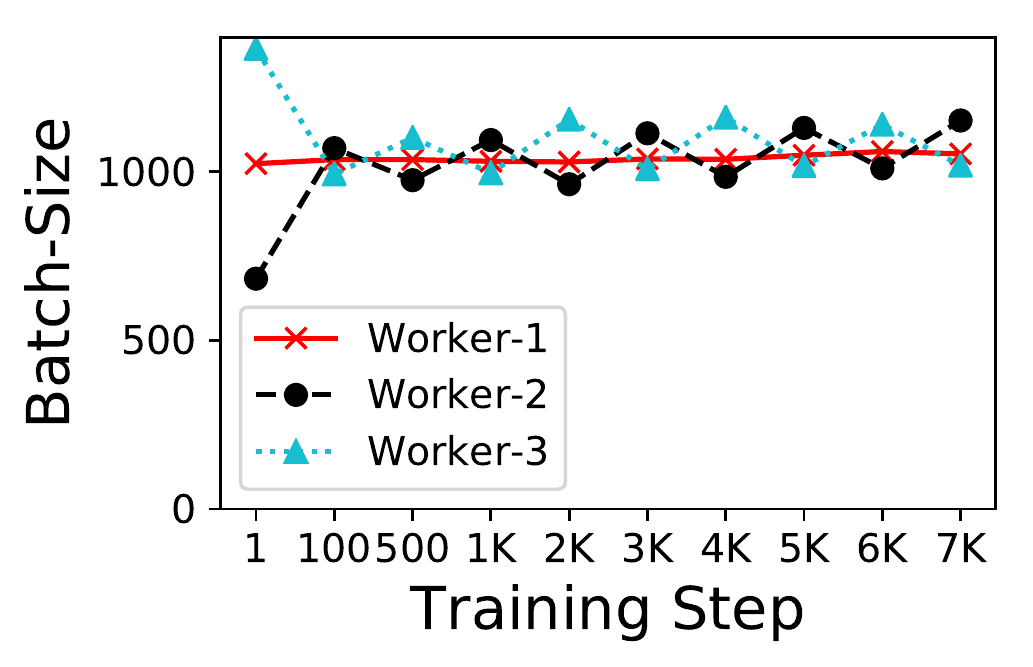} }
\caption{Dynamic batch size adjustments.}
\vspace*{\myfigspace}
\vspace*{\captionspace}
    \label{fig:ping-pong-all}
\end{figure}

The dynamic batch size adjustment can be done at the end of every iteration.
However, it is neither prudent nor necessary to do so.
%
Changing the batch size on workers is \emph{not} a zero-cost operation, because it involves terminating and restarting the training.
Furthermore, due to the stochastic nature of training, iteration times on workers will never converge to the exact average, and there will always be some error which the proportional control mechanism will try to chase.
This is illustrated in Figure~\ref{fig:ping-pong}, which shows the mini-batch sizes ``oscillating'' due to the dynamic batching adjustments. 


To prevent these oscillations and reduce the overhead of batch adjustments, we use three main techniques: dead-banding, exponential smoothing iteration times, and lower-upper bounds on batch sizes. We describe these approaches below:

\noindent \textbf{Dead-banding.}
After every iteration, we compute the new batch sizes using the proportional-control technique as described so far. 
We use a dead-band for our controller: batch sizes are updated only if the change is substantial.
We compute the difference between $b^{i+1} - b^i$ and do not update if this is smaller than threshold, $\Delta_{min}(b)$. 
If the change in the batch sizes on all workers is less than $\Delta_{min}(b)$, then no batch readjustment is made. 
The threshold can be chosen based on how sensitive we want the adjustment to be, and it also depends on the performance overhead of readjusting the batch sizes. 
For instance, current ML frameworks such as TensorFlow do not support graceful dynamic adjustment of batch sizes and require terminating and restarting the entire training process, in which case a larger threshold is preferable.
Based on the TensorFlow overheads, we use a dead-band threshold of 0.05: meaning that the new batch sizes on all workers must be atleast change by 5\%.

\noindent \textbf{Exponential Smoothing.}
With dead-banding, we only need to make batch adjustments at the start of the training process and whenever the underlying resource availability of the workers changes due to resource over-commitment or preemption. 
To improve the controller stability and avoid spurious readjustments, we compute the error (deviation of iteration time from the cluster average) on multiple iterations. 
Specifically, the error is computed using an EWMA (Exponentially Weighted Moving Average) across all the iterations since the previous batch readjustment. 
This provides us with the ``Integrator'' component in the controller, and particularly useful to prevent outliers.

With the dead-banding, we don't update batches on every iteration, and the moving average is computed in the interval with no batch size updates. 
Assume that last batch update happened on iteration $j$, and the current iteration is $i$. 
We then compute the average of worker $k$: $\mu(k,i,j)$ = EWMA($t_k^i, t_k^{i-1},...t_k^j$).
The smoothed iteration times ($\mu$) are used in Equation~\ref{eq:dbatch1} to compute the error and the batch size update.

\begin{figure}[t]
  \centering
  \vspace*{\myfigspace}
  \subfloat[Throughput on GPU \label{fig:tput_GPU}]
  {\includegraphics[width=0.22\textwidth]{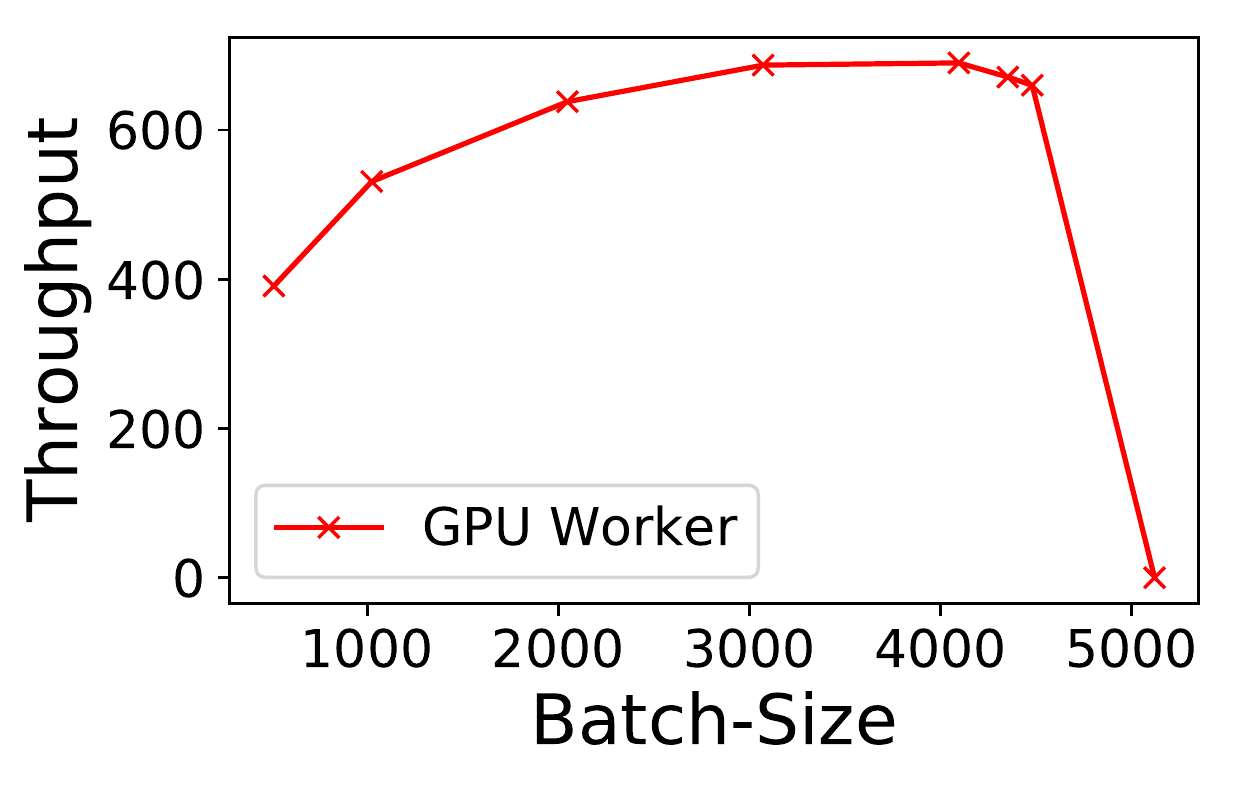}}
  \hfill 
  \subfloat[Throughput on CPU \label{fig:tput_CPU}]
  {\includegraphics[width=0.22\textwidth]{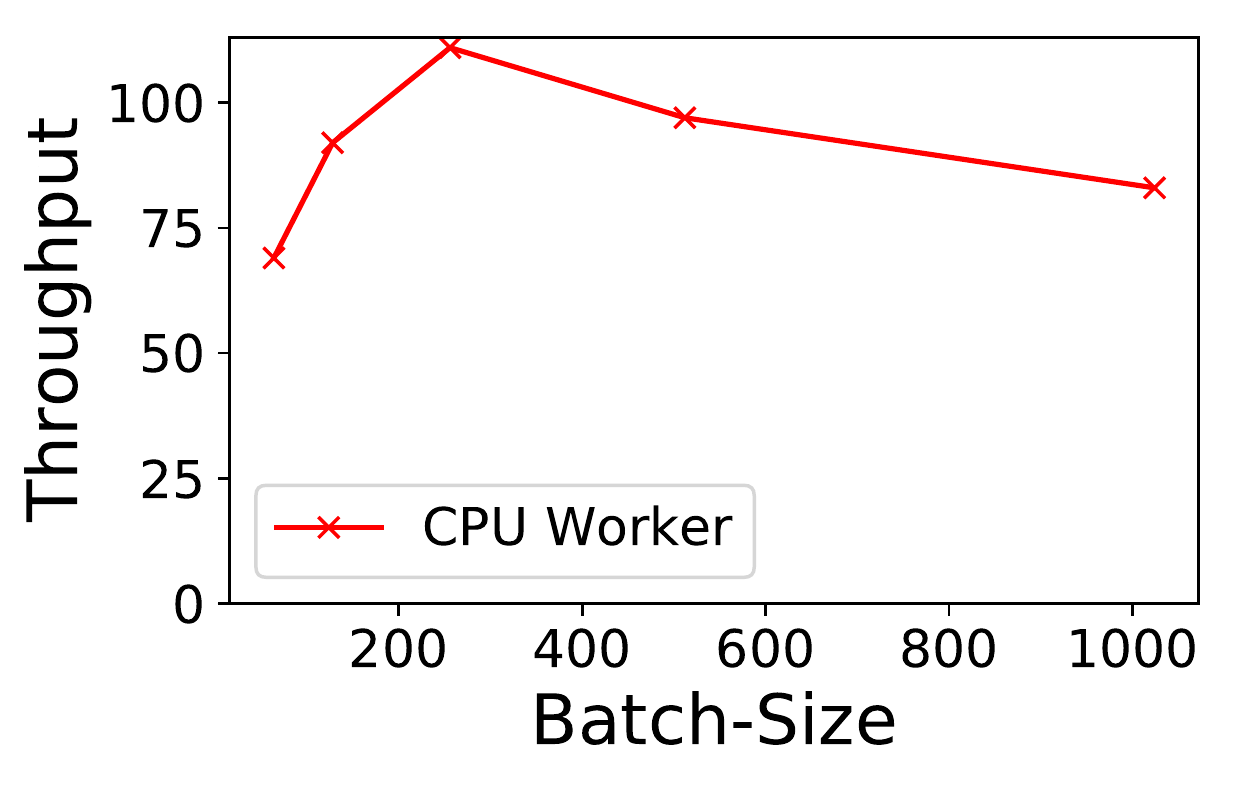}}
  \vspace*{\captionspace}
\caption{Training throughput (img/sec)  increases with batch size, then declines because of hitting resource (memory) limits on the workers, especially on GPUs where the memory limit is strict.}
\vspace*{\captionspace}
\label{fig:tput-b}
\end{figure}

\noindent \textbf{Batch size bounds.}
Finally, we enforce lower and upper bounds on mini-batch sizes on all workers.
These bounds prevent extreme batch sizes in cases of extreme heterogeneity, and ensure that the total throughput does not drop because of variable batching.
Extremely small batches cannot use all the hardware parallelism and yield low throughput.
Similarly, large batches may exhaust memory resources and also result in lower throughput.
This is illustrated in Figure~\ref{fig:tput-b}, which shows the throughput increasing with the batch size, until a sharp decline due to memory exhaustion in the GPU, and a gradual decline for CPU workers. 

We thus allow users to specify estimates of lower and upper bounds ($b_{\text{min}}, b_{\text{max}}$) of the batch sizes for all the workers.
As the training progresses and we readjust batch sizes, we get more data points for the throughput curve.
If we observe a drop in worker throughput after increasing its batch size from $b_0$ to $b$, then we update its $b_{\text{max}} = b_0$. This ensures that future batch readjustments will not result in a drop in throughput.

\noindent \textbf{Putting it all together.}
We can integrate all the control stability techniques into the proportional controller. Assume that the latest iteration is $i$, and the last batch-update was made in iteration $j$.
The pseudo-code for our dynamic batching can be expressed as:

\begin{enumerate}
\item Compute exponential moving average iteration times $\mu(k,i,j)$ for all workers $k$. 
\item Use $\mu(k,i,j)$ in Eqn~\ref{eq:dbatch1},~\ref{eq:dbatch2} to compute $\Delta(b_k)$ and $b_k^{i+1}$.
\item Enforce batch size bounds: $b_{k,min} \leq b_k \leq b_{k,max}$
\item Apply deadbanding check. If $\max_k{\Delta(b_k)/b_k} > \Delta_{min}(b)$, update all batch sizes. Otherwise do nothing. 
\end{enumerate}









\vspace*{\subsecspace}
\section{Experimental Evaluation} \label{sec:eval}
\vspace*{\subsecspace}

We conduct all our evaluation using our modified TensorFlow implementation that monitors differences in iteration times and dynamically adjusts per-worker batch sizes. 
We use the following standard well-known training workloads: 

\begin{itemize}[leftmargin=2pt]
\item \textbf{ResNet-50:} TensorFlow's ResNet benchmark~\cite{he2015deep}, trained on the standard CIFAR-10 dataset.
 We use a momentum optimizer with a learning rate schedule of [0.1, 0.01, 0.001, 0.0002]. 

\item \textbf{MNIST CNN~\cite{mnist}:} with  Adam~\cite{adam} and learning rate of 0.0001. 

\item \textbf{Linear Regression:}  To show our system effectively sustains heavy as well as comparatively lighter workloads, we perform Linear Regression (LR) on Harvard's bar crawl dataset \cite{Killian2019LearningTD}. 

\end{itemize}


\begin{figure*}[t]
  \centering
  \subfloat[ResNet-50  \label{fig:resnet-bsp}]
  {\includegraphics[width=0.3\textwidth]{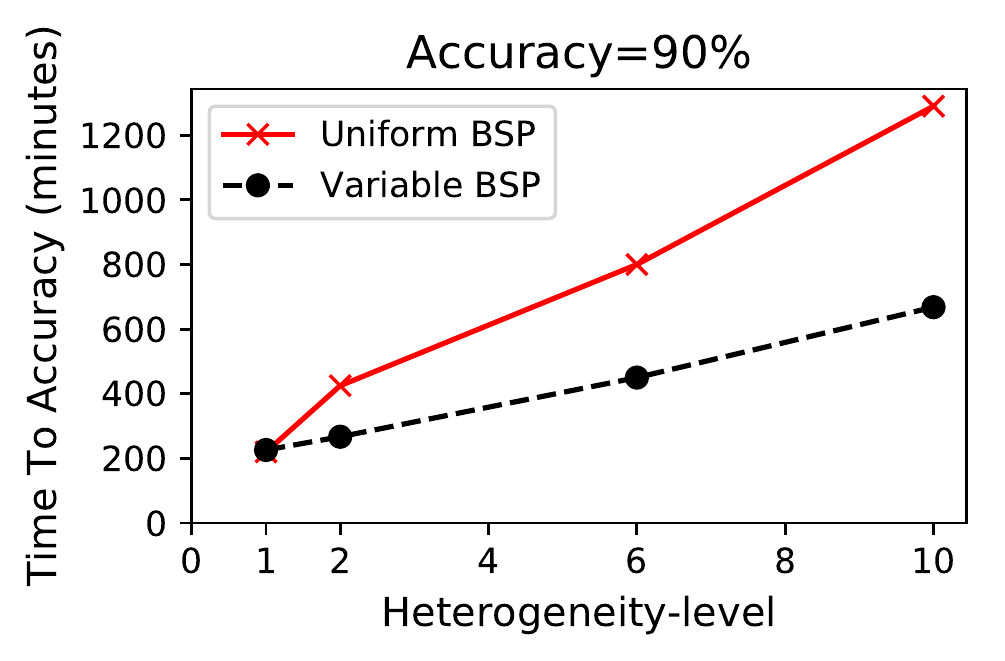}}
  \hfill
  \subfloat[MNIST CNN  \label{fig:mnist-bsp}]
  {\includegraphics[width=0.3\textwidth]{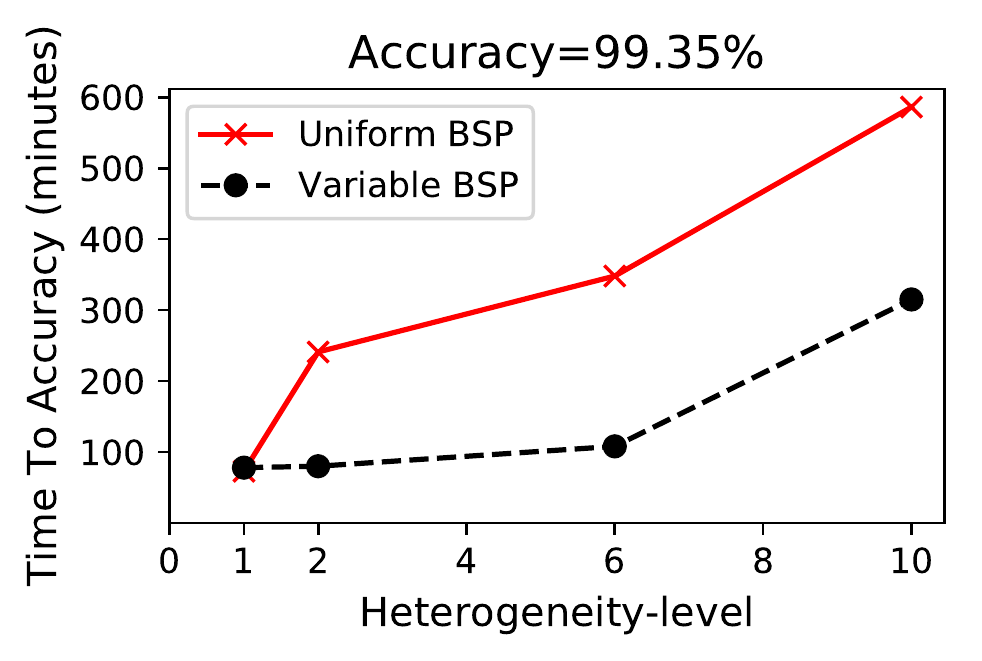}}
  \hfill
  \subfloat[Linear Regression  \label{fig:LR-bsp}]
  {\includegraphics[width=0.3\textwidth]{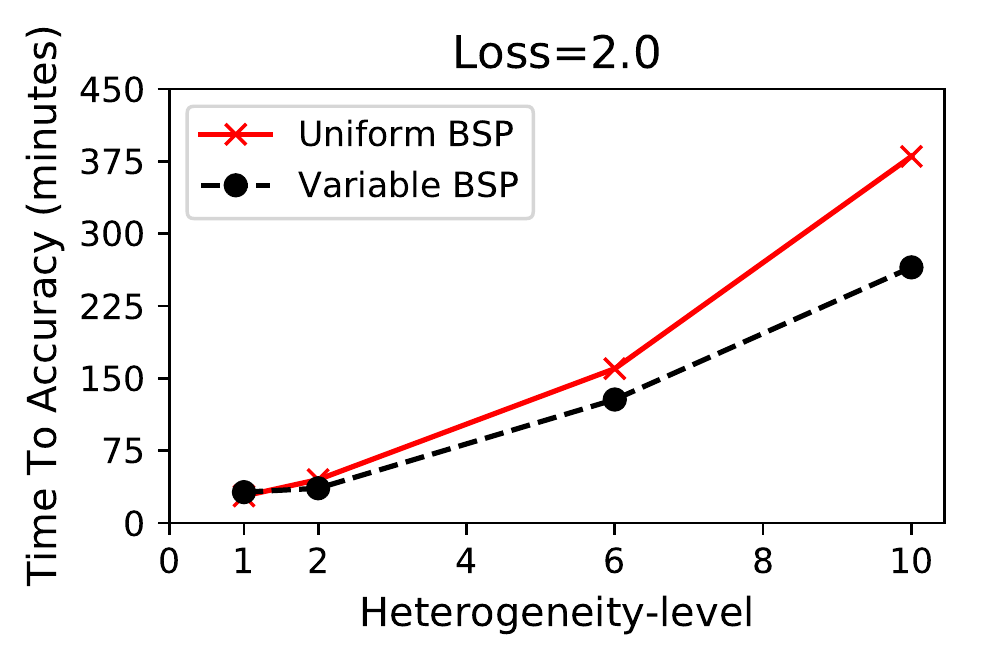}}
  \vspace*{\captionspace}
  \caption{With BSP synchronization, variable batching can reduce the total training time to accuracy by up to $4\times$.}
  \label{fig:bsp-all}
  \vspace*{\captionspace}
\end{figure*}

\noindent \textbf{Experimental environment and setup.}
We use the parameter server distribution strategy for all model training. 
We appropriately scale the number of parameter servers to ensure that they are not the bottleneck.
All TensorFlow processes (master, parameter servers, and workers) are deployed inside Docker containers for ease of management and fine-grained resource accounting and control.
We conduct all our empirical evaluation on a local cluster as well as on Google Cloud Platform.
The local cluster's CPU servers have 48-core Intel Xeon Platinum 2.10GHz CPUs and 256 GB of RAM. 
The GPU is Nvidia Tesla P100-PCIe-16GB. 



\vspace*{\subsecspace}
\subsection{CPU Training}
\vspace*{\subsecspace}

In this subsection, we focus on static heterogeneity when the cluster is composed of VMs/containers of different sizes. 
We are primarily interested in determining the impact of heterogeneity, and not parallel scaling. 
Therefore, we evaluate on clusters with different heterogeneity levels but the same total resource capacity. 
For instance, we compare a cluster configuration with two workers with (4, 16) CPUs, vs. two workers with (8, 12) CPUs. 
For CPU-only clusters, we define the  heterogeneity level as:
$  \text{H-level} = \max \text{number of cores}/\min \text{number of cores} $.

\noindent \textbf{Local cluster.}
We first present the training performance across different heterogeneity levels on our local cluster with three CPU workers.
The total number of CPU cores across the three workers is 39, and so a H-level of 2 would yield a (9, 12, 18) CPU cores configuration.
The total training time to reach a desired level of model accuracy across the three different workloads is shown in Figure~\ref{fig:bsp-all}.
Compared to vanilla TensorFlow's uniform batching, our variable batching approach can significantly reduce the training time. In general, the variable batching does better compared to the uniform batching at higher heterogeneity levels, because it is able to mitigate the stragglers.
For computationally intensive ResNet, our variable batching improves training times by $2\times$ at H-level of 2, and $2.4\times$ at the highest H-level of 10. 
%
%
The high heterogeneity levels result in very small workers (e.g., H-level 10 is a (2,17,20) configuration). 
The small workers end up being stragglers even with variable batching's load balancing, because we are not able to use any parallelism inside these small workers, yet still face the same communication and model synchronization overhead.

The MNIST CNN also sees a performance improvement of $2\times$---$4\times$. 
Finally, the Linear Regression workload is the least computationally expensive, and sees the least benefit ($\sim 15\%$) from the load-balancing that variable batching provides, because it is communication and synchronization bound. 

Importantly, our variable batching can ameliorate the heterogeneity-induced slowdown, and can ``flatten the curve''.
At a high H-level of 6, ResNet training time only increases by $2\times$ compared to the homogenous setup (Figure~\ref{fig:bsp-all}).
Similarly, MNIST time increases by by $4\times$, and Linear Regression by only $5\%$.

\noindent \textbf{Result:} \emph{Variable batching can mitigate stragglers in BSP and can reduce training time by $4\times$ for high heterogeneity levels. Our technique is particularly effective in scenarios that are computation and not communication bound.}

\vspace*{\subsecspace}
\vspace*{\subsecspace}
\subsection{GPU Training}
\vspace*{\subsecspace}

\begin{figure}[H]
  \centering
  \vspace*{\myfigspace}
  \subfloat[GPU-CPU mix.   \label{fig:gpu-cpu}]
  {\includegraphics[width=0.22\textwidth]{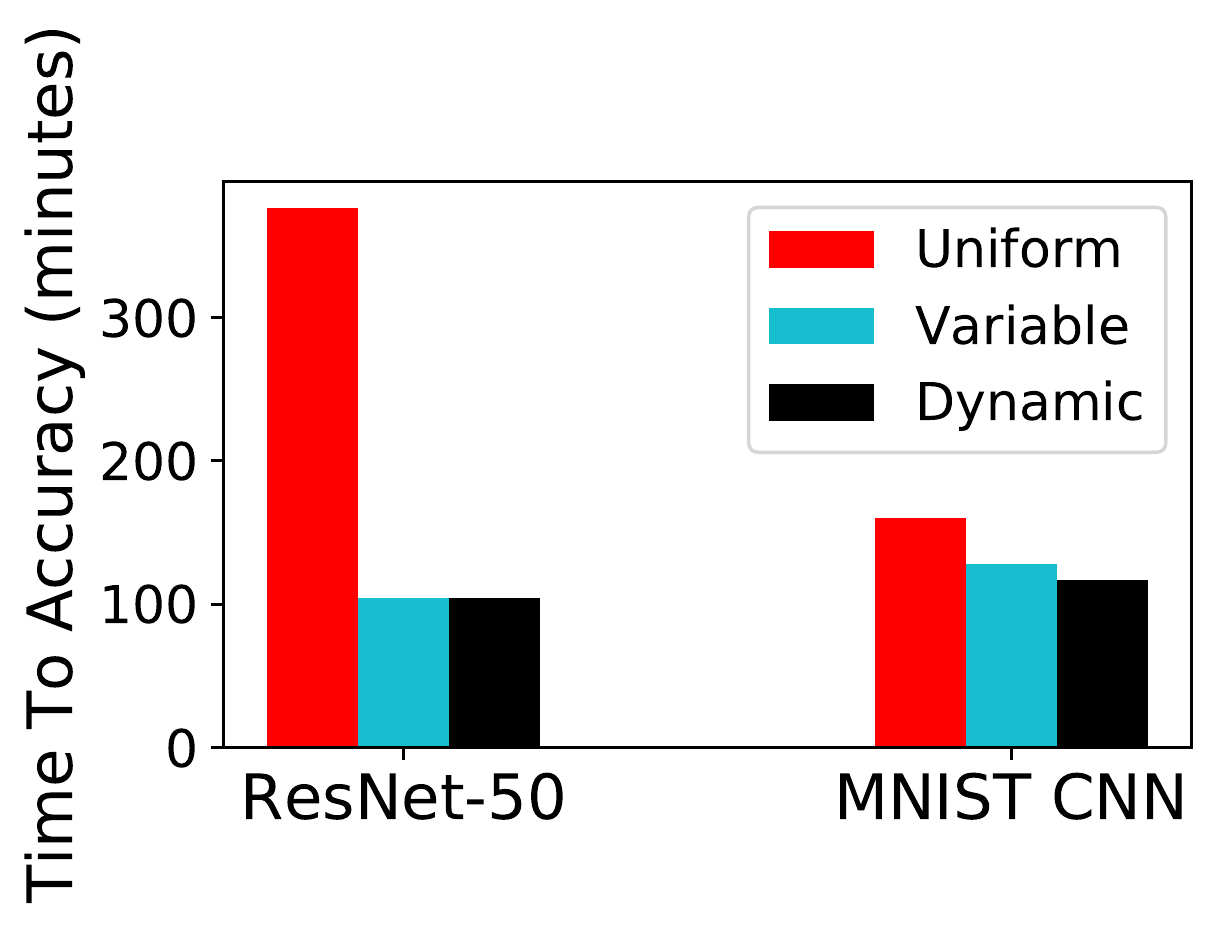}}
  \hfill
  \subfloat[ResNet with multiple GPU types. \label{fig:multi-gpu}]
  {  \includegraphics[width=0.22\textwidth]{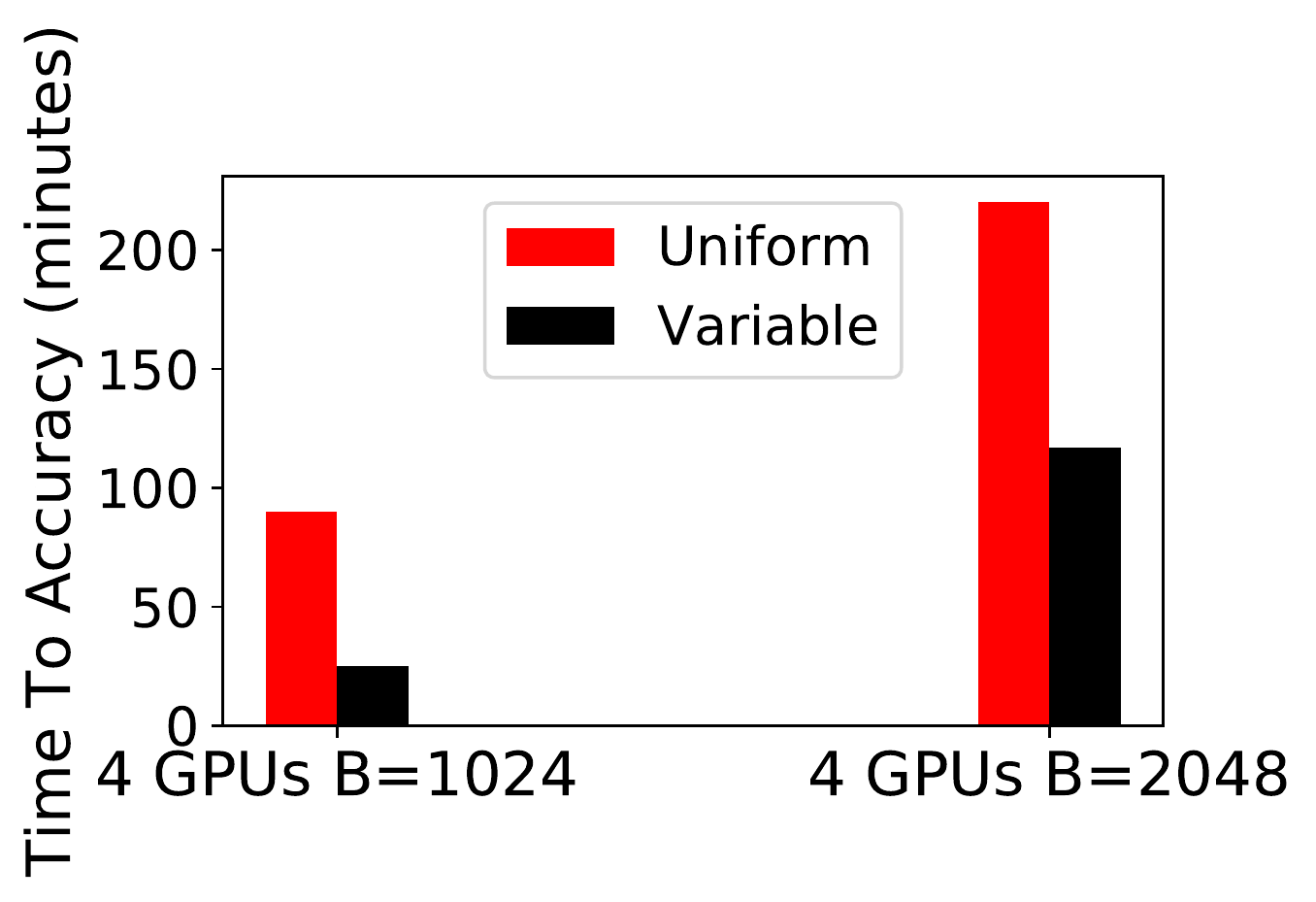}}
  \vspace*{\captionspace}
  \label{fig:GPU}
  \caption{GPU training.}
    \vspace*{\captionspace}
\end{figure}

For GPU training, we first consider an extreme heterogeneity case where the cluster comprises of both CPU and GPU workers.
Specifically, we use a single GPU worker (Tesla P100) and CPU worker (48-core Intel Xeon).
We compare the performance of uniform, variable, and dynamic batching in Figure~\ref{fig:gpu-cpu}.

Recall that variable batch allocation is an open-loop approach that assigns batch sizes based on the hardware FLOPs performance and not actual throughput.
Compared to uniform batching, we are able to reduce the training time by more than $4\times$ for the computationally intensive ResNet workload. 
For MNIST, the cluster is underutilized, since workload is not computationally bound, and we see a more modest 20\% improvement in training time with our approach.

The performance of the Xeon Platinum CPUs used in our local cluster experiments is far closer to GPU performance most cloud CPUs. 
For instance, the ratios of the FLOPs and the batch size between the GPU and CPU was $0.813:0.187$, and thus the GPU worker is ``only'' $4.3\times$ faster.

Interestingly, the dynamic batching improves performance by about 3\% compared to static variable batching for MNIST CNN, and has a negligible effect for ResNet.
This intriguing result is because of the tradeoff of dynamic batching.
For a computationally intensive workload like ResNet, hardware FLOPs approximates throughput, so there was not enough opportunity for the dynamic readjustments.
The kill-restart approach poses a small performance overhead too.
These two factors ``cancel out'' and in most cases, static variable batching is ``good enough''.

\noindent \textbf{Result:} \emph{Variable batching allows efficient use of mixed GPU-CPU clusters, and can reduce the training time by up to  $4\times$.}

We also examine training performance on a cloud cluster with two different types of GPUs.
Specifically, we run two VMs with Tesla T4 and two VMs with Tesla P4 GPUs.
The training time of ResNet (BSP) was 90 minutes with uniform batching, and only 20 minutes with variable batching----a $4.5\times$ improvement.

\vspace*{\subsecspace}
\section{Related Work}
\vspace*{\subsecspace}
\label{sec:related}




\noindent \textbf{Heterogeneous Training.}
The closest work is~\cite{jiang_heterogeneity-aware_2017}, which develops synchronization techniques (DynSGD and ConSGD), for mitigating the effects of staleness and stragglers by explicitly accounting for staleness using a vector-clock technique. 
However much like other work in this area~\cite{luo_hop:_2019}, the cluster heterogeneity they consider is only a result of stochastic performance variations (random worker slowdowns).
Instead, we focus on \emph{systemic} and severe heterogeneity due to vastly different resource sizes of workers.  
Our fundamental idea of variable mini-batch sizes is agnostic to the synchronization technique and can also be integrated with ConsSGD to provide support for alleviating the random slowdowns due to performance interference. 


Heterogeneity in training is being recognized as an important missing feature and many approaches are being developed. 
~\cite{wang_heterogeneity-aware_2019} uses a gradient coding scheme to tolerate stragglers due to  static heterogeneity in a BSP setup.
Our variable batching technique is applicable in existing parameter server based architectures and does not require gradient coding. 
Heterogeneity for decentralized training is explored in Hop~\cite{luo_hop:_2019}, which uses a bounded staleness approach and bound the iteration-gap. 
The technique is shown to be effective in case of random worker slowdowns. 
Its effectiveness at high static heterogeneity levels is less clear, since the large iteration gaps may pose fundamental synchronization challenges in the decentralized setting.

Resource allocation for training is also an active area of work, and is challenging due to our incomplete first-principles understanding of SGD scaling, and profiling-driven empirical models are typically used. 
~\cite{peng_optimus_2018} shows how to do cluster resource allocation and scheduling for ML training jobs by developing and using an empirical performance model to determine number of workers and parameter servers to use.
Similarly, Cynthia~\cite{zheng_cynthia:_2019} uses an analytical performance model for cost efficient cloud resource provisioning. 
In contrast, our approach can directly start training without the need for apriori modeling.
Our design goal was to design a generally usable mechanism  that is plug-in compatible with different resource allocation approaches, training algorithms, and treats ML models as ``black boxes''.
Integrated systems and training algorithm co-design, like in Orpheus~\cite{xie_orpheus:_2018} that improves consistency via periodic centralized synchronization, is an alternative approach. 



\noindent \textbf{Model synchronization} impacts training performance, especially in cloud environments with higher stochasticity in server performance and network latencies. 
This has motivated many synchronization techniques such as stale synchronous parallel~\cite{ho_more_ssp} and others~\cite{cui2014bounded-staleness, harlap2016addressing, zhao_sync---fly:_2018, zhang_stay_2018, wei_managed_2015}.
The performance tradeoffs of synchronization techniques in dynamic cloud environments is studied in~\cite{ambati_understanding_2019}.
Although asynchronous approaches~\cite{recht2011hogwild} seem promising in heterogeneous environments, gradient staleness is still a pernicious problem
~\cite{chen_revisiting_2016, dutta_slow_2018, zhou_distributed_nodate, haddadpour_local_nodate}. 



\noindent \textbf{Batch size}
in distributed training is one of the most crucial hyper-parameters that affects the training performance as well as the model convergence. 
Understanding these tradeoffs is a key problem in machine learning ~\cite{shallue_measuring_2018, yin_small_2017,zhou_distributed_nodate,gupta_model_2016}.
%
%
Due to the duality between learning rates and global batch sizes~\cite{smith_bayesian_2018}, 
adjusting the \emph{global} batch size is a known technique to regulate the errors in SGD training~\cite{yu_computation_2019}.
Adabatch~\cite{devarakonda_adabatch_2018} and~\cite{smith_dont_2018} describe a ``batch size schedule'' analogous to a learning rate schedule. 
This is distinct from our dynamic mini-batch adjustment, and the dynamic global batch schedules can easily be incorporated into our approach.
Finally, the theoretical soundness of variable mini-batch sizes can be found in~\cite{ferdinand_anytime_2019}. 
They also propose a new synchronization technique where gradient updates are ``pulled'' from workers periodically, irrespective of their mini-batch processing, resulting in different sized worker updates. 





\noindent \textbf{Acknowledgments.} This work was partially supported by the Google Cloud research credits program.







\vspace*{\subsecspace}



{
\bibliographystyle{IEEEtran}
\interlinepenalty=10000 
\bibliography{ml,scicloud,spot-bid}
}

\end{document}